\documentclass[conference]{IEEEtran}
\IEEEoverridecommandlockouts
\usepackage{cite}
\usepackage{amsmath,amssymb,amsfonts}
\usepackage{multirow}
\usepackage{amsfonts}
\usepackage{booktabs}
\usepackage{subfigure}
\usepackage{amsmath}
\usepackage{algorithmic}
\usepackage{graphicx}
\usepackage{textcomp}
\usepackage{xcolor}
\def\BibTeX{{\rm B\kern-.05em{\sc i\kern-.025em b}\kern-.08em
    T\kern-.1667em\lower.7ex\hbox{E}\kern-.125emX}}
\begin{document}

\title{FashionFAE: Fine-grained Attributes Enhanced\\
Fashion Vision-Language Pre-training
\thanks{$^{*}$These authors contributed equally to this work.$^\dag$Corresponding author. © 2025 IEEE.  Personal use of this material is permitted. Permission from IEEE must be obtained for all other uses, in any current or future media, including reprinting/republishing this material for advertising or promotional purposes, creating new collective works, for resale or redistribution to servers or lists, or reuse of any copyrighted component of this work in other works.}}
\author{
    \IEEEauthorblockN{
        Jiale Huang$^{1*}$, 
        Dehong Gao$^{3,4*}$, 
        Jinxia Zhang$^{1,2\dag}$, 
        Zechao Zhan$^{1}$,
        Yang Hu$^{1}$,
        Xin Wang$^{5}$
    }
    \IEEEauthorblockA{
        $^{1}$\textit{Key Laboratory of Measurement and Control of CSE, School of Automation, Southeast University, China}\\
        $^{2}$\textit{Advanced Ocean Institute of Southeast University, Nantong, China}
        $^{3}$\textit{Northwestern Polytechnical University, China}\\
         $^{4}$\textit{Binjiang Institute of Artificial Intelligence, ZJUT, Hangzhou, China}
        $^{5}$\textit{Alibaba Group, China}\\
    }
}
\maketitle

\begin{abstract}
Large-scale \textbf{V}ision-\textbf{L}anguage \textbf{P}re-training (VLP) has demonstrated remarkable success in the general domain.
However, in the fashion domain, items are distinguished by fine-grained attributes such as texture and material, which are crucial for tasks such as retrieval. Existing models often fail to take advantage of these fine-grained attributes from both text and image modalities.
To address the above issue, we propose a novel approach for the fashion domain, \textbf{F}ine-grained \textbf{A}ttributes \textbf{E}nhanced VLP (FashionFAE), which focuses on the detailed characteristics of the fashion data. An attribute-emphasized text prediction task is proposed to predict fine-grained attributes of the items.
This forces the model to focus on the salient attributes from the text modality.
In addition, a novel attribute-promoted image reconstruction task is proposed, which further enhances the fine-grained ability of the model by leveraging the representative attributes from the image modality.
Extensive experiments show that FashionFAE outperforms \textbf{S}tate-\textbf{O}f-\textbf{T}he-\textbf{A}rt (SOTA) methods, achieving 2.9\% and 5.2\% improvements in retrieval on sub-test set and full test set, respectively, and an average improvement of 1.6\% in recognition tasks.
\end{abstract}


\section{Introduction}

Vision-Language Pre-training (VLP) has experienced remarkable progress \cite{li2019visualbert,su2019vl,li2020oscar,kim2021vilt} in recent years, becoming a fundamental cornerstone for various multimodal tasks. VLP models learn general knowledge from numerous image-text pairs, enabling them to handle various downstream tasks effectively.
\par
Numerous studies have attempted to apply VLP models to fashion-related tasks \cite{lu2019vilbert,li2021align,wu2024self}. 
\begin{figure}[htbp]
    \centering
    \includegraphics[width=0.7\linewidth]{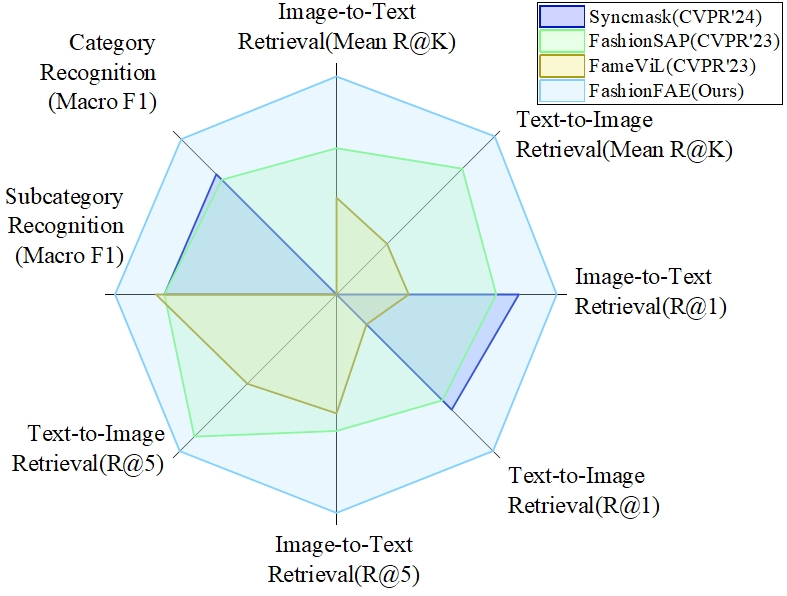}
\caption{FashionFAE achieves SOTA performance in various metrics for cross-modal retrieval and (sub)category recognition in the fashion domain.}
\label{SOTA}
\end{figure}
However, directly using general pre-trained models does not yield optimal results, as they lack specific adaptability to the nuances of fashion data. Compared to the general domain, the fashion domain relies more on \textbf{fine-grained information}, such as materials and textures, to distinguish items, rather than simply focusing on coarse-grained information, such as category and location. To enhance the suitability of VLP models for the fashion domain, attention to fine-grained information is essential. Efforts are already underway to focus on fine-grained information in the fashion domain \cite{han2022fashionvil,han2023fashionsap}. Nevertheless, these models still encounter the following challenges:

\par
1) \emph{In the fashion domain, besides primary descriptive information, there are additional attributes that contain unique features and significant information. Existing research has not explicitly modeled these crucial attributes.} 
\par
2)\emph{Current models overlook detailed visual attributes by treating all image parts equally. Focusing only on refining fine-grained text representations is insufficient. A genuine and comprehensive understanding requires the acquisition of fine-grained information from both text and images.}
\begin{figure*}[htbp]
	\centering
	\begin{minipage}{0.9\linewidth}
		\centering
		\includegraphics[width=1\linewidth]{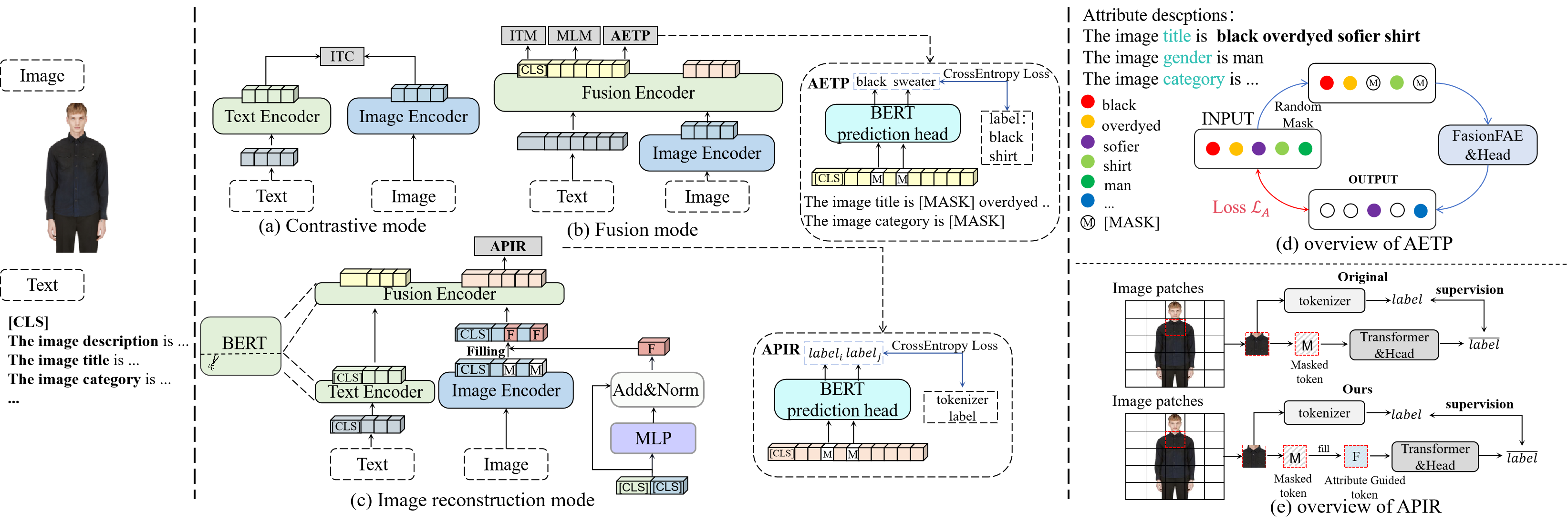}
	\end{minipage}
 \caption{Overview of our FashionFAE model architecture and proposed AETP and APIR tasks. To accommodate the different 5 pre-training tasks, the model has a total of 3 modes: (a) Contrastive mode; (b) Fusion mode; (c) Image reconstruction mode.}
 \label{model overview}
\end{figure*}
\par
To solve these problems, we propose a novel approach for the \textbf{fashion} domain, called \textbf{F}ine-grained \textbf{A}ttributes \textbf{E}nhanced VLP (FashionFAE), which simultaneously extracts fine-grained information from both the textual and visual sides within the fashion domain. Specifically, our approach introduces fine-grained information into our FashionFAE model through the Attribute-Emphasized Text Prediction (AETP) task on the textual side and the Attribute-Promoted Image Reconstruction (APIR) task on both visual and textual sides. It achieves superior performance on a set of diverse fashion tasks, as in Fig.~\ref{SOTA}. The overview of the proposed AETP and APIR tasks is illustrated in Fig.~\ref{model overview} (d) and (e). 

The AETP task is designed to fully leverage the wealth of additional attribute information available within the fashion dataset. These attributes encompass a variety of elements, including item title, category, subcategory, season, composition, and gender. Firstly, attributes and their respective values are structured into complete attribute description statements and incorporated into the input text. Then, the attribute values are randomly masked and predicted. The AETP task not only maximizes the utilization of various attributes in the fashion dataset but also empowers the model to discern and extract intricate attribute details, contributing to a more comprehensive and insightful analysis of the data.
\par
The APIR task is proposed for both visual and textual modalities. For an input image, it is divided into a grid of patches, which are then discretized into visual labels using latent codes from a Vector Quantized Variational Autoencoder (VQ-VAE). During the APIR task, certain image patches are masked, and their corresponding labels are predicted.
Auxiliary information from the text assists in predicting attributes in missing image patches, enhancing its ability to capture fine-grained attribute details.
\par
Our main contributions are summarized below: 1)An effective fashion vision-language pre-training model is proposed, which delves deep into the characteristics of fashion domain data; 2)We propose the AETP task, enhancing the model's ability to capture salient attribute information; 3)We propose the APIR task, enabling the model to more comprehensively learn fine-grained attributes within the images; 4)FashionFAE achieves a new SOTA with a significant improvement in image-to-text retrieval, text-to-image retrieval and (sub)category recognition.
\section{Method}
\subsection{Model Overview}
FashionFAE's architecture, shown in Fig.~\ref{model overview}, includes an image encoder based on Vision Transformer \cite{dosovitskiy2020image} and a transformer module based on BERT \cite{devlin2018bert}. 
\par
To achieve a fashion pre-training model with high performance, the architecture comprises three modes that can flexibly handle our proposed pre-training tasks, AETP and APIR, as well as three common pre-training tasks. For different modes, the transformer module based on BERT can function as either a text encoder or a fusion encoder, or it can be divided into two separate components, with one acting as the text encoder and the other as the fusion encoder. Importantly, all these modules share identical parameters.
\par
For the input text, a primary description as well as additional attributes are employed to fully leverage the fine-grained information of the item. From the FashionGen dataset, a total of six crucial attributes are identified: \textit{title}, \textit{category}, \textit{subcategory}, \textit{gender}, \textit{composition}, and \textit{season}. For each attribute and its corresponding value, a comprehensive attribute statement is constructed in the following format: ``\textit{The image} [\textit{attribute}] \textit{is} [\textit{value}]." (e.g., ``\textit{The image category is shirts.}"). 
All attribute statements are appended following the initial description, collectively constituting the textual input provided to the model. 

\subsection{Pre-training Tasks}
\subsubsection{Attribute-Emphasized Text Prediction (AETP)}
In attributes like title and category, rich fine-grained information is embedded. These attributes elevate the uniqueness of items, contributing to improved retrieval and classification tasks. The introduction of the AETP task guides the model in concentrating on noteworthy features in the text, thereby augmenting its capacity to comprehend fine-grained details and construct more discriminative representations of items. In this task, an item's title consists of adjectives or nouns describing it (e.g., \textit{title: Black Overdyed Sofier Denim Shirt}), with each word as a subattribute. Other attributes like category and subcategory each align with a single attribute value (e.g., \textit{category: shirts}).
We randomly sample $N$ subattributes from the title along with one from the remaining five attributes. The values of the chosen attributes are replaced with a special token [MASK].
\par
The masked text input and the image embedding are fed into the fusion encoder, where the prediction head predicts the attributes that are masked with the remaining text embedding and image embedding.
With the image-text pair $(v,w)$ sampled from the dataset $D$, the AETP task is optimized by minimizing the negative log-likelihood:
\begin{equation}\label{AETP}
    \mathcal{L}_{AETP} = -\mathbb{E}_{(v,w) \sim D} \log{P_{\theta}(a_{m}\vert w_{\backslash m},v)}
\end{equation}
where $P$ represents the predicted probability, $\theta$ represents the learnable parameters, $a_{m}$ represents masked attributes, $w_{\backslash m}$ represents the embedding without masked attributes $m$ and $v$ represents the image embedding.
\begin{figure}[htbp]
    \centering
    \includegraphics[width=0.7\linewidth]{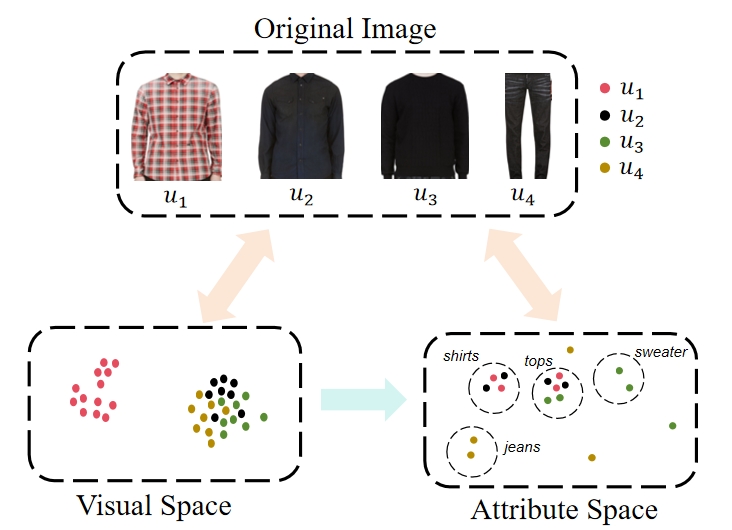}
\caption{The visual representations of black shirts, black sweaters, and black pants are indistinguishable by showing similarities in the visual space but can be well differentiated in more advanced attribute space.}
\label{space}
\end{figure}

\subsubsection{Attribute-Promoted Image Reconstruction (APIR)}
In the fashion domain, images often resemble each other in patches due to shared characteristics such as color, texture, and material. As illustrated in Fig.~\ref{space}, relying solely on pixel-level semantics proves to be insufficient in capturing the entirety of an item's characteristics. Therefore, it is necessary to use a more sophisticated attribute space, particularly at the attribute level, to effectively convey an item's defining attributes. To better abstract embeddings into the semantic space at the attribute level, reconstructing the attributes of these patches is more appropriate. Incorporating characteristics from the textual domain also improves image reconstruction. Thus, the APIR task is proposed to enhance the image modality's capacity to capture detailed information by incorporating prior knowledge from the text modality. The goal of this task is to predict the patch label provided by an offline image tokenizer pre-trained on the fashion dataset.
\par

We randomly replace the 25\% of the image patches with the fusion token $F$, which is generated based on the token $w_{cls}$ from the text encoder and $v_{cls}$ from the image encoder.
\begin{equation}\label{fusion token}
    F = LN(MLP(v_{cls}+w_{cls})+v_{cls}+w_{cls})
\end{equation}
\par
The model can be trained to predict the tokenized label of each masked patch by minimizing the negative log-likelihood.
\begin{equation}\label{APIR}
    \mathcal{L}_{APIR} = -\mathbb{E}_{(v,w) \sim D} \log{P_{\theta}(v_{m}^{t}\vert v_{\backslash m},F,w)}
\end{equation}
where $v_{m}^{t}$ represents the target label for the masked patch, $v_{\backslash m}$ represents the image embedding of unmasked image patches, $w$ represents the text embedding.

\subsubsection{Image-Text Contrastive Learning (ITC)}
The ITC task is used to promote close representations of the image and the corresponding text in latent space. 
To assess the similarity between the given text embedding and the image embedding$(v,w)$, the dot product of their average pooling feature is utilized as a metric. To enhance the similarity, a symmetrical contrastive loss function is employed:

\begin{equation}\label{ITC}
    \mathcal{L}_{ITC} = \frac{1}{2}[\mathcal{L}_{InfoNCE}(v,w) + \mathcal{L}_{InfoNCE}(w,v)]
\end{equation}
\begin{equation}\label{InfoNCE}
    \mathcal{L}_{InfoNCE} = -\mathbb{E}_{(x,y) \sim B}\log{\frac{exp(s(x,y))}{\sum_{\tilde{y} \in B}exp(s(x,\tilde{y}))}}
\end{equation}
where $s(\cdot,\cdot)$ represents similarity calculation, $(x,y)$ represents the matched pair, $(x,\tilde{y})$ represents the unmatched pair within a batch $B$.

\subsubsection{Masked Language Modeling (MLM)}
Conventional Masked Language Modeling (MLM) is used to model and learn from the description.  We use a prediction head with parameters shared with the AETP task. MLM is optimized by minimizing the negative log-likelihood:

\begin{equation}\label{MLM}
    \mathcal{L}_{MLM} = -\mathbb{E}_{(v,w) \sim D} \log{P_{\theta}(w_{m}^{t}\vert w_{\backslash m},v)}
\end{equation}
where $w_{m}^{t}$ represents the target label for masked subwords, $w_{\backslash m}$ represents the embedding of unmasked subwords.
\subsubsection{Image-Text Matching (ITM)}
In the ITM task, a binary label $L$ is predicted by employing the [CLS] token outputted from the fusion mode, which indicates if each input image-text pair is a match. We apply the cross-entropy loss for the ITM task:
\begin{equation}\label{ITM}
    \mathcal{L}_{ITM} = -\mathbb{E}_{(v,w) \sim T} \log{P_{\theta}(L\vert (v,w))}
\end{equation}
\par
The complete pre-training objective is the combination of the mentioned tasks above, we randomly sample one task per iteration with the sampling probability $P_{task}$.
\begin{equation}\label{ALL}
\nonumber
\mathcal{L} = \mathcal{L}_{task}
\end{equation}
\begin{equation}\label{ALL1}
task \in {\{AETP,APIR,ITC,MLM,ITM\}}_{P_{task}}
\end{equation}

\section{Experiments}

\subsection{Dataset}
The FashionGen dataset \cite{rostamzadeh2018fashion} is used for pre-training and downstream tasks. FashionGen includes 292k text-image pairs and 60k unique fashion items (260k for training and 32k for evaluation). A detailed description of the item is provided in the text annotation as well as a variety of attributes. Downstream tasks, i.e., text-to-image retrieval, image-to-text retrieval, category recognition, and subcategory recognition, are evaluated based on the test set of the FashionGen dataset.

\subsection{Implementation Details}
The proposed FashionFAE is implemented using Multi-Modal Framework (MMF) \cite{singh2020mmf} and PyTorch \cite{paszke2019pytorch}. The hyperparameter $N$ (the number of masked subattributes in the title) of the proposed AETP task is set to 2. An AdamW optimizer with a learning rate $1e$-$5$ is adopted. The batch size is 128. During the pre-training stage, the MLM task and the AETP task are optimized simultaneously. 

\subsection{Downstream Tasks and Results}
\subsubsection{Cross-modal Retrieval}
We address both image-to-text retrieval and text-to-image retrieval on FashionGen dataset using two protocols employed by previous methods: (1) Random 100 Protocol (sub-test set): For each query, 100 candidates from the same category are randomly sampled to form a retrieval database \cite{gao2020fashionbert,zhuge2021kaleido}. (2) Full Database Protocol (full test set): We also embrace a more challenging and practical protocol that involves retrieval across the entire product set. Evaluation metrics include $Rank$@$K, \{K=1,5,10\}$ and their mean value. 
\par
\begin{table}[!htbp] 
    \centering
    \caption{Cross-modal retrieval results with the sub-test set.}
        \resizebox{1\linewidth}{!}{
            \begin{tabular}{ccccccccccc}
            \toprule
            \multirow{2}*{Methods}& \multirow{2}*{Pub./Year} & \multicolumn{4}{c}{I2T} & \multicolumn{4}{c}{T2I} &\multirow{2}*{Mean}\\
            \cmidrule(lr){3-6}\cmidrule(lr){7-10}
             & & R@1 & R@5 & R@10 & Mean & R@1 & R@5 & R@10 & Mean &\\
            \midrule
            FashionBERT \cite{gao2020fashionbert} & SIGIR$_{20}$ &23.96 &46.31 &52.12 &40.80 &26.75 &46.48 &55.74 &42.99 &41.89\\
            KaleidoBERT \cite{zhuge2021kaleido} & CVPR$_{22}$ &27.99 &60.09 &68.37 &52.15 &33.88 &60.60 &68.59 &54.36 &53.25\\
            CommerceMM \cite{yu2022commercemm} & SIGKDD$_{22}$ &41.60 &64.00 &72.80 &59.47 &39.60 &61.50 &72.70 &57.93 &58.70  \\
            EI-CLIP \cite{ma2022ei} & CVPR$_{22}$ &38.70 &72.20 &84.25 &65.05 &40.06 &71.99 &82.90 &64.98 &65.02\\
            MLVT \cite{ji2023masked} & MIR$_{23}$ &33.10 &77.20 &91.10 &67.13 &34.60 &78.00 &89.50 &67.37 &67.25\\
            FashionKLIP \cite{wang2023fashionklip}& ACL$_{23}$ &60.79 &85.67 &91.95 &79.47 &54.00 &78.49 &86.28 &72.92 &76.20\\
            FaD-VLP \cite{mirchandani2022fad} & EMNLP$_{22}$ &64.30 &86.75 &93.48 &81.51 &58.66 &84.92 &91.58 &78.39 &79.95 \\
            FashionViL \cite{han2022fashionvil}& ECCV$_{22}$ &65.54 &91.34 &96.30 &84.39 &61.88 &87.32 &93.22 &80.81 &82.60\\
            Fame-ViL \cite{han2023fame}& CVPR$_{23}$ &65.94 &91.92 &97.22 &85.03 &62.86 &87.38 &93.52 &81.25 &83.14\\
            FashionSAP \cite{han2023fashionsap} & CVPR$_{23}$ &73.14&92.80 &96.87 &87.60 &70.12 &91.76 &96.39 & 86.09 &86.85\\
            SyncMask \cite{song2024syncmask} & CVPR$_{24}$ &75.00&- &- &- &71.00 &- &- & - &-\\
            \midrule
            Ours & - &\textbf{78.12} & \textbf{96.88} & \textbf{98.98} &\textbf{91.33} & \textbf{74.95} & \textbf{92.96} & \textbf{96.60} &\textbf{88.17} &\textbf{89.75}\\
            \bottomrule
        \end{tabular}
        }
  \label{subset}
\end{table}
\begin{table}[!htbp] 
    \centering
    \caption{Cross-modal retrieval results with full test set.}
        \resizebox{1\linewidth}{!}{
            \begin{tabular}{ccccccccccc}
            \toprule
            \multirow{2}*{Methods} &\multirow{2}*{Pub./Year}& \multicolumn{4}{c}{I2T} & \multicolumn{4}{c}{T2I} &\multirow{2}*{Mean}\\
            \cmidrule(lr){3-6}\cmidrule(lr){7-10}
             & & R@1 & R@5 & R@10 &Mean  & R@1 & R@5 & R@10 &Mean &\\
            \midrule
            EI-CLIP \cite{ma2022ei} & CVPR$_{22}$ &25.70 &54.50 &66.80 &49.00 &28.41 &57.10 &69.40 &51.64 &50.32\\
            FashionKLIP \cite{wang2023fashionklip}& ACL$_{23}$ &37.01 &59.78 &67.39 &54.73 &43.70 &63.74 &72.67 &60.04 &57.38\\
            FashionViL \cite{han2022fashionvil}&ECCV$_{22}$ &42.88 &71.57 &80.55 &65.00 &51.34 &75.42 &84.57 &70.44 &67.72\\
            FashionSAP \cite{han2023fashionsap}&CVPR$_{23}$ &54.43&77.30 &83.15 &71.63 &62.82 &83.96 &90.16 &78.98 &75.31\\
            SyncMask \cite{song2024syncmask} &CVPR$_{24}$&55.39&- &- &- &64.06 &- &- & - &-\\
            \midrule
            Ours & - &\textbf{57.51} & \textbf{83.81} & \textbf{89.75} &\textbf{77.02} & \textbf{68.34} & \textbf{89.10}  &\textbf{94.51} &\textbf{83.98} &\textbf{80.50}\\

            \bottomrule
        \end{tabular}
        }
 \label{fullset}
\end{table}
The results in Tab.~\ref{subset} show that our method significantly outperforms the latest SOTA approach in all metrics for I2T and T2I tasks on the sub-test set, with an average improvement of 3.5\% in the R@1 metric and an overall performance gain of 2.9\%. On the more challenging full test set, as shown in Tab.~\ref{fullset}, our model also excels, achieving a 5.2\% improvement in mean retrieval performance and a 3.2\% increase in R@1 accuracy compared to previous SOTA models.
\par

\subsubsection{Category/Subcategory Recognition (CR\&SCR)}
Following the previous works \cite{zhuge2021kaleido,han2022fashionvil}, the CR\&SCR tasks are evaluated on the FashionGen dataset. Evaluation metrics include Accuracy (Acc) and Macro-Averaged F1 Score (Macro-F1). The results are shown in Tab.~\ref{CR and SCR}. FashionFAE achieves the highest Acc and also significantly improves the Macro-F1 score (2.6\% on average) in both CR\&SCR tasks.
\begin{table}[!htbp]
    \centering
    \caption{CR and SCR results on FashionGen.}
    \resizebox{0.8\linewidth}{!}{
        \begin{tabular}{ccccccc}
        \toprule
        \multirow{2}*{Methods}&\multirow{2}*{Pub./Year} & \multicolumn{2}{c}{CR} & \multicolumn{2}{c}{SCR} &\multirow{2}*{Mean}\\
            \cmidrule(lr){3-4}\cmidrule(lr){5-6}
        & &Acc & Macro-F1 & Acc & Macro-F1 &\\
        \midrule
        FashionViL\cite{han2022fashionvil} &ECCV$_{22}$ &97.48 &88.60 &89.23 &83.02 &89.58\\
        Fame-ViL \cite{han2023fame} &CVPR$_{23}$ &- &- &94.67 &88.21 &91.44\\
        FashionSAP \cite{han2023fashionsap} &CVPR$_{23}$ &98.34 &89.84 &94.33 &87.67 &92.55\\
        SyncMask \cite{song2024syncmask} &CVPR$_{24}$ &98.41 &90.31 &94.21 &87.83 &92.69\\
        \midrule
        Ours & - & \textbf{98.79} &\textbf{93.32} &\textbf{95.25} &\textbf{90.07} &\textbf{94.36}\\
        \bottomrule
    \end{tabular}
    }
    \label{CR and SCR}
\end{table}

\subsection{Ablation Study}
The effectiveness of the pre-training tasks adopted is analyzed in this section. The average metrics on the sub-test set and full test set for both I2T and T2I tasks, as well as the average value of Acc and Macro-F1 for CR \& SCR, are listed. The results of the ablation study are shown in Tab.~\ref{ablation}. The results of ablation experiments for ITC, ITM, and MLM, widely used in various VLP models, are depicted in the first three rows.
\par
\begin{table}[!htbp] 
    \centering
    \caption{Ablation study results for adopted pre-training tasks.}
        \resizebox{1\linewidth}{!}{
            \begin{tabular}{ccccccccccccc}
            \toprule
            \multirow{2}*{ITC}&\multirow{2}*{ITM}&\multirow{2}*{MLM}&\multirow{2}*{AETP} & \multirow{2}*{APIR}& \multicolumn{3}{c}{sub-test set} & \multicolumn{3}{c}{full test set} &CR &SCR\\
            \cmidrule(lr){6-8}\cmidrule(lr){9-11} \cmidrule(lr){12-12} \cmidrule(lr){13-13}
              &&&&  & I2T & T2I & Mean & I2T  & T2I & Mean & Mean  & Mean\\
            \midrule
            \checkmark & & & & & 82.82 &78.04 &80.43 &63.18 & 69.68 &66.43 &91.17 &84.78\\
            \checkmark&\checkmark & & && 82.71 &78.52 &80.61 &63.48 & 69.80 &66.64 &92.46 &86.04\\
             \checkmark &\checkmark &\checkmark & &&  83.23&79.10 &81.17 &63.70 & 70.45 &67.07 &92.21 &85.65\\
             \checkmark &\checkmark &\checkmark &\checkmark &&  90.84 &87.45 &89.14 &76.03 & 83.44 &79.74 &95.73 &91.68\\
            \checkmark&\checkmark & \checkmark & \checkmark &\checkmark& \textbf{91.33} &\textbf{88.17} &\textbf{89.75} &\textbf{77.02} & \textbf{83.98} &\textbf{80.50} &\textbf{96.05} &\textbf{94.69}\\
            \bottomrule
        \end{tabular}
        }

    \label{ablation}
\end{table}
The last two rows of the table present ablation experiments for AETP and APIR. Compared to more widely adopted methods, AETP and APIR are more focused on exploring the features of fashion domain data, enabling the model to pay a thorough attention to the representative characteristics of the data. The ablation experiments demonstrate that the AETP task has highlighted the crucial attributes inherent to items, substantially enhancing their discriminative qualities. This enhancement has led to significant performance improvements across various downstream tasks. The APIR task further exploits the attribute information present in the image domain and synergizes with the textual information, resulting in additional enhancements beyond the already commendable performance achieved.
\par

To validate the effectiveness of individual attributes (\textit{title}(TI), \textit{category} (CA), \textit{subcategory} (S-CA), \textit{gender} (GEN), \textit{composition} (COM), \textit{season} (SEA)) in the AETP task, we conducted ablation experiments. As shown in Tab.~\ref{AETP ablation}, the first row represents the scenario without additional attributes. The 2nd to 7th rows progressively add one attribute at a time, showing the effectiveness of each attribute.

\begin{table}[!htbp]
    \centering
    \caption{Ablation study results for AETP task.}
   \resizebox{0.8\linewidth}{!}{
        \begin{tabular}{cccccccccc}
        \toprule
        \multirow{2}*{TI}&\multirow{2}*{CA}& \multirow{2}*{S-CA} & \multirow{2}*{GEN}& \multirow{2}*{SEA}& \multirow{2}*{COM}&\multicolumn{2}{c}{sub-test set} & \multicolumn{2}{c}{full test set}\\
            \cmidrule(lr){7-8}\cmidrule(lr){9-10}
        &  & &  &&  &I2T & T2I & I2T  & T2I\\
        \midrule
        & && && & 83.23 &79.11 &63.70&70.45 \\
        \checkmark& && && & 83.75 &79.67 &64.64 &70.73\\
        \checkmark&\checkmark && && & 83.73 &80.10 &65.20 &71.38\\
        \checkmark& \checkmark&\checkmark& && & 83.59 &79.39 &65.44 &71.30\\
        \checkmark&\checkmark &\checkmark&\checkmark && & 85.68 &81.53 &67.84 &74.02\\
        \checkmark&\checkmark &\checkmark&\checkmark &\checkmark& & 91.25 &87.69 &76.75 &83.47\\
        \checkmark&\checkmark &\checkmark&\checkmark &\checkmark&\checkmark & \textbf{91.33} &\textbf{88.17} &\textbf{77.02} &\textbf{83.98}\\
        \bottomrule
    \end{tabular}
    }
    \label{AETP ablation}
\end{table}

\section{Conclusion}
In this paper, we propose FashionFAE, a fine-grained attribute enhanced vision-language pre-training model for the fashion domain. The model includes attribute-emphasized text prediction and attribute-promoted image reconstruction tasks to extract fine-grained information from both text and images. Comparisons and ablation studies show that FashionFAE significantly outperforms SOTA models, with a 2.9\% improvement in average retrieval on the sub-test set, a 5.2\% improvement on the full test set, and a 1.8\% improvement in category and subcategory recognition.
\vfill\pagebreak

\bibliographystyle{IEEEtran}
\bibliography{ref}{}

\end{document}